\documentclass[letterpaper,journal]{IEEEtran}

\usepackage{amsmath,amsfonts,amssymb}
\usepackage{array}
\usepackage[caption=false,font=normalsize,labelfont=sf,textfont=sf]{subfig}
\usepackage{textcomp}
\usepackage{stfloats}
\usepackage{url}
\usepackage{verbatim}
\usepackage{graphicx}
\usepackage{cite}
\usepackage{booktabs}
\usepackage{multirow}
\usepackage{bm}
\usepackage{balance}
\usepackage{xcolor}
\usepackage{makecell}
\usepackage{tabularx}

\allowdisplaybreaks[1]
\renewcommand{\arraystretch}{0.94}
\newcommand{\SE}{\mathrm{SE}(3)}
\newcommand{\SO}{\mathrm{SO}(3)}
\newcommand{\R}{\mathbb{R}}
\newcommand{\Tobj}{\mathcal{T}_{obj}}

\newcommand{\Freach}{\mathcal{F}_{reach}}
\newcommand{\Smask}{\mathbf{S}}
\newcommand{\SPD}{\mathcal{S}^{++}}

\newcommand{\dairm}{d_{\mathrm{AIRM}}}

\begin{document}

\title{PhysV2A: Reachability-Gated and Semantic-Mask-Constrained Feasibility Completion for Video-to-Robot Manipulation}

\author{Haohui Huang,
Junda Duan,
Tao Teng,
and Chenguang Yang%
\thanks{Haohui Huang and Junda Duan are with the School of Automation, Guangdong University of Technology, Guangzhou, China (e-mail: hh.huang@ieee.org; 2112404293@mail2.gdut.edu.cn).}%
\thanks{Tao Teng is with the University of Liverpool, Liverpool, U.K. (e-mail: Tao.Teng@liverpool.ac.uk).}%
\thanks{Chenguang Yang is with the Department of Computing, The Hong Kong Polytechnic University, Hong Kong SAR, China (e-mail: cyang@ieee.org).}%
\thanks{Corresponding author: Chenguang Yang.}%
}

\markboth{Preprint}{Huang \MakeLowercase{\textit{et al.}}: PhysV2A}

\maketitle

\begin{abstract}
Video-based manipulation provides object-centric motion priors from human demonstrations, generated videos, or RGB-D observations, but such priors are typically embodiment-agnostic and cannot be directly executed by a specific robot. This paper presents \textbf{PhysV2A}, a reachability-gated and semantic-mask-constrained feasibility-completion framework for converting video-derived 6D object motion into robot-executable manipulation trajectories. The key idea is to treat grasp feasibility as trajectory-conditioned rather than local: each RGB-D-generated 6-DoF grasp candidate is rigidly coupled with the recovered object motion to form a grasp-conditioned TCP trajectory hypothesis. PhysV2A then performs hierarchical reachability-gated selection, where infeasible grasp--trajectory pairs are rejected by robot-centric kinematic checks and surviving candidates are ranked by downstream execution suitability. For the selected reachable trajectory, a VLM-assisted and rule-validated S-Mask identifies task-critical and relaxable Cartesian components, enabling semantic-mask-constrained manipulability refinement through redundancy-first optimization and bounded Cartesian relaxation. Real-robot experiments on four tabletop manipulation tasks show that PhysV2A improves task success over representative video-prior and IK-only baselines, reduces kinematic-feasibility failures, and produces better-conditioned trajectories with bounded semantic deviations.
\end{abstract}


\section{Introduction}
\IEEEPARstart{R}{ecent} advances in video generation, RGB-D tracking, vision-language reasoning, and robot learning have made visual observations an increasingly useful source of task-level priors for robot manipulation. Human demonstrations, generated videos, and RGB-D observations can describe how objects should move during manipulation, while point tracking and 6D pose recovery provide geometric interfaces for converting image-space motion into object-space trajectories \cite{doersch2023tapir,karaev2023cotracker,wen2024foundationpose}. Video-conditioned manipulation methods further exploit observed or generated videos as object-centric demonstrations, where the visual sequence specifies task intent without directly specifying robot joint commands \cite{li2025novaflow,patel2025rigvid,bharadhwaj2025gen2act,dharmarajan2025dream2flow}. These developments motivate a video-to-robot manipulation pipeline that recovers object motion from visual priors and retargets it to a specific robot embodiment.

\begin{figure}[!t]
\centering
\includegraphics[width=0.94\columnwidth]{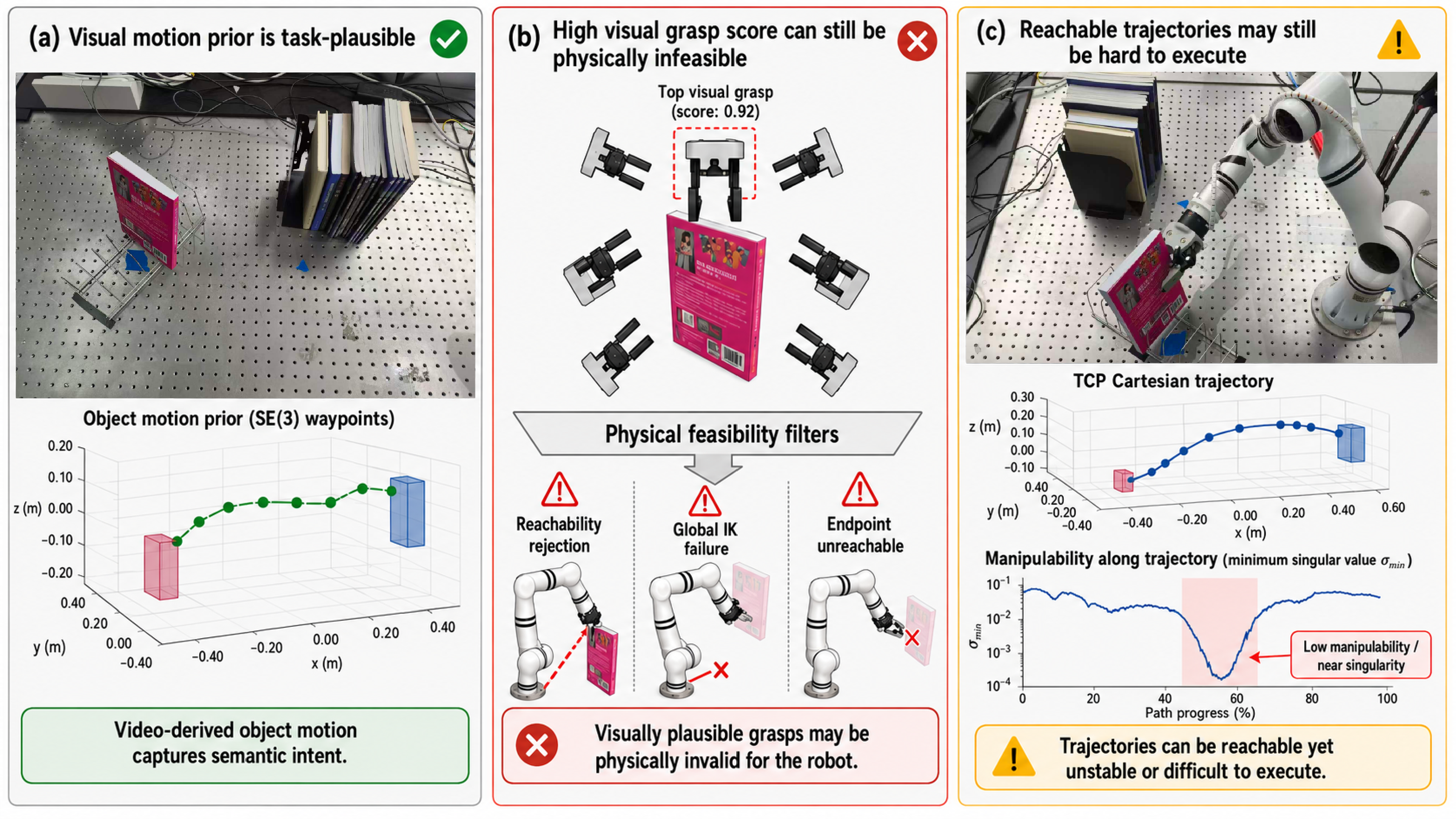}
\caption{\textbf{Motivation.} Video-derived motion priors can encode task intent, but visually plausible grasps and trajectories may still violate reachability, full-trajectory IK, endpoint constraints, or manipulability requirements.}
\label{fig:motivation}
\end{figure}

However, visual motion plausibility does not directly imply robot-specific executability. A video-derived object trajectory may be coherent in image or object space, but the TCP trajectory induced on a specific manipulator can still violate workspace limits, joint limits, IK continuity, or singularity-related constraints. More importantly, grasp feasibility in video-to-robot manipulation is not a purely local property of a grasp pose. A high-confidence 6-DoF grasp candidate generated from RGB-D observations may be locally plausible in the point cloud \cite{bohg2014data,fang2020graspnet,sundermeyer2021contact,fang2023anygrasp,murali2025graspgen}, yet may induce an unreachable or poorly conditioned TCP trajectory after being rigidly coupled with the recovered object motion. This motivates a trajectory-conditioned view of grasp feasibility: a grasp should be evaluated by the full downstream TCP trajectory it induces, rather than by local grasp confidence alone.

A second challenge is that feasibility cannot be reliably determined by a single endpoint check or by binary IK validation alone. Candidate grasp--trajectory pairs may fail at different levels, including workspace support, start or terminal IK, full-trajectory IK consistency, joint-limit margin, joint-space continuity, and manipulability safety. Therefore, video-to-robot retargeting requires a hierarchical reachability-gated selection mechanism that first rejects infeasible candidates through hard robot-centric constraints and then ranks the remaining candidates according to execution suitability.

A third challenge is that even a reachable trajectory may be poorly conditioned for execution. Improving manipulability by freely perturbing the Cartesian trajectory can reduce singularity risk, but it may also distort task-critical motion components such as placement position, insertion direction, or object lift-off. Thus, manipulability refinement should be semantic-mask-constrained: task-critical Cartesian components should be preserved, while only semantically relaxable components should be adjusted within explicit bounds.

We propose \textbf{PhysV2A}, a robot-centric feasibility-completion framework built around three coupled components. First, PhysV2A formulates grasp feasibility as a trajectory-conditioned problem by rigidly coupling each 6-DoF grasp candidate with the recovered object motion to generate a grasp-conditioned TCP trajectory. Second, it performs hierarchical reachability-gated selection, rejecting infeasible grasp--trajectory pairs before refinement and ranking feasible candidates by downstream execution suitability. Third, it refines the selected trajectory through semantic-mask-constrained manipulability optimization, where a VLM-assisted and rule-validated S-Mask constrains Cartesian relaxation to task-preserving motion components while SPD manipulability geometry provides the execution-quality objective. In this work, feasibility refers to pre-execution kinematic feasibility and manipulability safety rather than complete collision-aware planning, contact-rich dynamics, or closed-loop force control.

In summary, the contributions of this work are as follows:
\begin{itemize}
\item We formulate \textbf{trajectory-conditioned grasp feasibility} for video-to-robot manipulation by evaluating each RGB-D grasp proposal through the complete TCP trajectory it induces after being coupled with video-derived object motion.
\item We propose a \textbf{hierarchical reachability-gated selection} strategy that combines hard robot-centric feasibility screening with soft execution-suitability ranking over workspace support, QGMM reachability, trajectory-level IK consistency, joint-limit margin, joint continuity, and manipulability-related conditioning.
\item We develop a \textbf{semantic-mask-constrained manipulability refinement} module in which a VLM-assisted and rule-validated S-Mask constrains redundancy-first SPD manipulability refinement and bounded Cartesian relaxation without distorting task-critical motion components.
\end{itemize}
Real-robot experiments on four tabletop manipulation tasks validate that the proposed formulation improves task success over representative video-prior and IK-only same-platform baselines while producing better-conditioned trajectories with bounded semantic deviations.

\begin{figure*}[!t]
    \centering
    \includegraphics[width=0.95\textwidth]{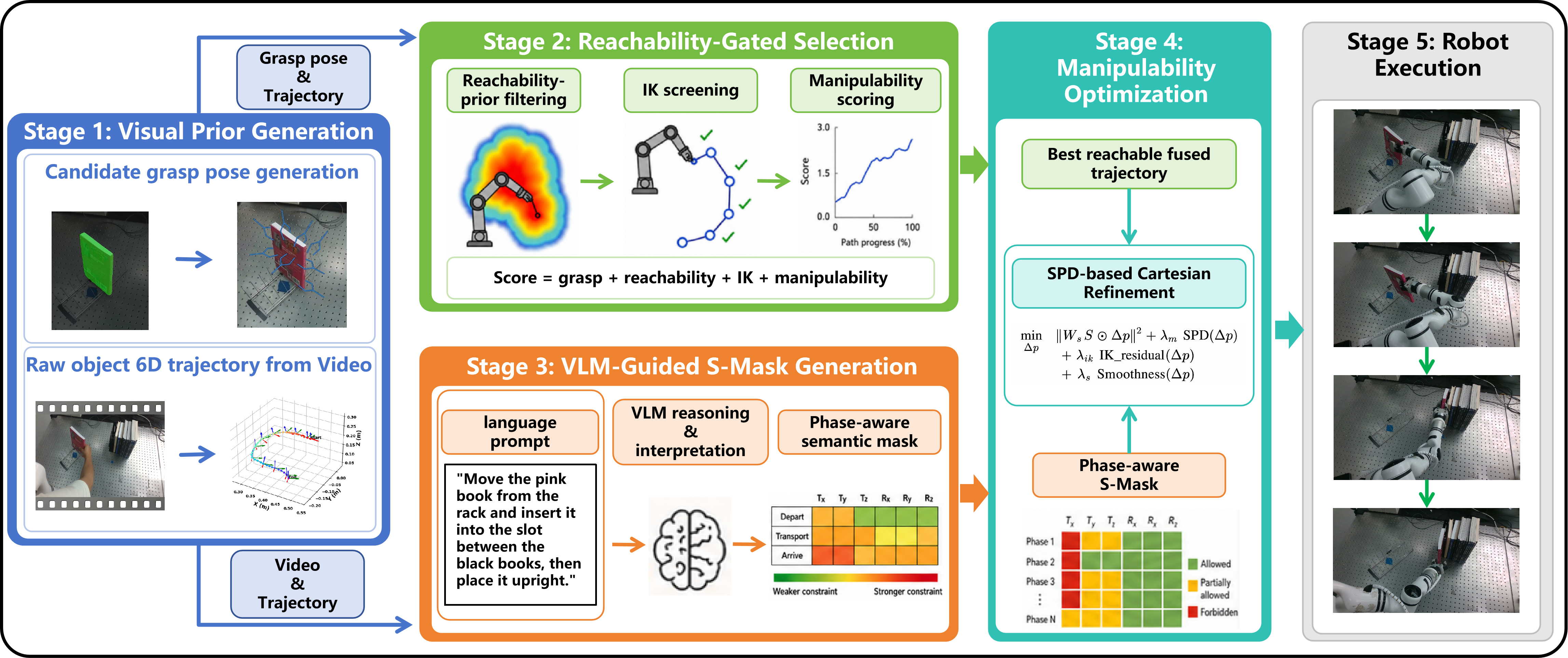}
    \caption{\textbf{Overall framework.} PhysV2A converts video-derived object motion and RGB-D grasp proposals into executable robot trajectories. It constructs grasp-conditioned TCP hypotheses, selects reachable candidates through robot-centric screening and ranking, generates a phase-aware S-Mask from visual and trajectory cues, and refines the selected trajectory through redundancy-first and S-Mask-constrained manipulability optimization.}
    \label{fig:framework}
\vspace{-0.3em}
\end{figure*}

\section{Related Work}

\subsection{Visual Action Priors for Robot Manipulation}
Vision-based manipulation systems use keypoints, dense flow, 6D poses, affordance fields, and generated action videos to encode task intent. Generalist robot policies and vision-language-action models can predict actions when embodiment-specific robot data are available \cite{brohan2022rt1,octo2024octo,chi2023diffusion,zitkovich2023rt2}. Video-conditioned methods use observed or generated videos as object-level priors rather than direct joint commands \cite{li2025novaflow,patel2025rigvid,bharadhwaj2025gen2act,dharmarajan2025dream2flow}, while tracking and 6D pose estimation provide geometric interfaces for recovering object-space motion \cite{eisner2022flowbot3d,doersch2023tapir,karaev2023cotracker,wen2024foundationpose}. These outputs remain embodiment-agnostic and do not guarantee reachability, IK continuity, or manipulability on a specific robot.

\subsection{RGB-D Grasp Generation and Grasp-Conditioned Feasibility}
Modern 6-DoF grasp-generation methods predict dense parallel-jaw grasp candidates from partial RGB-D observations or point clouds \cite{bohg2014data,fang2020graspnet,sundermeyer2021contact,fang2023anygrasp,murali2025graspgen}. However, grasp confidence is usually a local geometric measure. A grasp that is plausible in the point cloud may still induce an unreachable or poorly conditioned TCP trajectory after being coupled with video-derived object motion. PhysV2A therefore treats grasp generation as a proposal stage and evaluates each grasp by the downstream trajectory it induces.

\subsection{Reachability, Inverse Kinematics, and Manipulability}
Reachability modeling, capability maps, and constrained-manipulability maps represent whether an end-effector pose is achievable under workspace and kinematic constraints \cite{vahrenkamp2015workspace}. Probabilistic orientation models can represent pose distributions on quaternion spaces \cite{kim2017gaussian}, and redundant-manipulator IK and manipulability measures based on singular values, condition numbers, and SPD geometry are widely used for feasibility analysis and trajectory optimization \cite{buss2004introduction,siciliano1990kinematic,nakamura1986inverse,yoshikawa1985manipulability,jaquier2021geometry}. PhysV2A applies these robot-centric measures during grasp-conditioned trajectory selection and refinement.

\subsection{VLMs as Semantic Constraint Generators}
Vision-language models have been used for instruction grounding, task planning, robot program synthesis, value-map generation, and spatial constraint generation for manipulation \cite{ahn2022saycan,driess2023palme,zitkovich2023rt2,liang2023code,singh2023progprompt,huang2023voxposer,huang2025rekep}. These semantic priors do not by themselves guarantee robot-specific feasibility. PhysV2A therefore uses the VLM only to generate the S-Mask, while reachability checking, IK validation, and manipulability refinement remain grounded in robot kinematics.

\section{Problem Formulation}

\subsection{Input Representation and Trajectory-Conditioned Grasp Feasibility}

The upstream visual module provides an object-centric 6D trajectory
\begin{equation}
    \Tobj=\{T_{obj}^{t}\}_{t=0}^{N},
    \qquad
    T_{obj}^{t}=(R_{obj}^{t},p_{obj}^{t})\in\SE ,
    \label{eq:obj_traj_def}
\end{equation}
where each pose is expressed in the robot base frame. The RGB-D-based grasp generation module provides a set of candidate TCP grasp poses
\begin{equation}
    \mathcal{G}=\{(G_i,s_i^{grasp})\}_{i=1}^{K},
    \qquad
    G_i=(R_{G_i},p_{G_i})\in\SE ,
    \label{eq:grasp_set_def}
\end{equation}
where $s_i^{grasp}$ denotes the local grasp confidence.

For each candidate grasp, the initial object-to-TCP attachment is defined as
\begin{equation}
    T^{0}_{obj\rightarrow tcp,i}
    =
    (T_{obj}^{0})^{-1}G_i .
    \label{eq:object_to_tcp}
\end{equation}
Assuming that the object remains rigidly attached to the gripper after grasping, the grasp-conditioned TCP trajectory induced by $G_i$ is
\begin{equation}
    T_{tcp,i}^{t}
    =
    T_{obj}^{t}T^{0}_{obj\rightarrow tcp,i}
    =
    T_{obj}^{t}(T_{obj}^{0})^{-1}G_i .
    \label{eq:tcp_rigid_fusion}
\end{equation}
Equivalently, its translational and rotational components are
\begin{equation}
\begin{aligned}
    p_{tcp,i}^{t}
    &=
    p_{obj}^{t}
    +
    R_{obj}^{t}(R_{obj}^{0})^{T}
    (p_{G_i}-p_{obj}^{0}),\\
    R_{tcp,i}^{t}
    &=
    R_{obj}^{t}(R_{obj}^{0})^{T}R_{G_i}.
\end{aligned}
\label{eq:pos_fusion}
\end{equation}
Thus, one video-derived object trajectory induces multiple grasp-conditioned TCP trajectories,
\begin{equation}
    \mathcal{T}_{tcp}^{(i)}
    =
    \{T_{tcp,i}^{t}\}_{t=0}^{N}.
    \label{eq:tcp_traj_i}
\end{equation}

\subsection{Reachable Feasible Set}

Let $q_t\in\R^n$ denote the robot joint configuration, $FK(q_t)=(R(q_t),p(q_t))$ the forward kinematics, and $J(q_t)$ the geometric Jacobian. To avoid mixing translational and rotational units in Jacobian-based metrics, we use the task-space scaled Jacobian
\begin{equation}
    J_s(q_t)=W_xJ(q_t),
    \qquad
    W_x=\mathrm{diag}(l_0^{-1}I_3,I_3),
    \label{eq:scaled_jacobian}
\end{equation}
where $l_0$ is a fixed characteristic length used consistently for all methods and tasks, and the Jacobian is ordered as translational rows followed by rotational rows. A grasp-conditioned trajectory $\mathcal{T}_{tcp}^{(i)}$ belongs to the reachable feasible set $\Freach$ only if there exists a continuous joint sequence $\{q_t\}_{t=0}^{N}$ satisfying
\begin{equation}
\begin{aligned}
    &q_{min}\le q_t\le q_{max},\\
    &\|p(q_t)-p_{tcp,i}^{t}\|_2\le\epsilon_p,\\
    &d_R(R(q_t),R_{tcp,i}^{t})\le\epsilon_R,\\
    &m_{limit}(q_t)\ge\epsilon_{limit},\\
    &\sigma_{\min}(J_s(q_t))\ge\sigma_{safe},\\
    &\kappa(J_s(q_t))\le\kappa_{safe},
    \quad t=0,\ldots,N,\\
    &\|q_t-q_{t-1}\|_2\le\epsilon_{\Delta q},
    \quad t=1,\ldots,N .
\end{aligned}
\label{eq:reach_constraints}
\end{equation}
Here $d_R(\cdot,\cdot)$ is the geodesic orientation error on $\SO$, and $m_{limit}(\cdot)$ is the normalized joint-limit margin:
\begin{equation}
    d_R(R_1,R_2)
    =
    \|\log(R_1^{T}R_2)\|_2,
    \label{eq:rotation_error}
\end{equation}
\begin{equation}
    m_{limit}(q_t)
    =
    \min_r
    \frac{
    \min(q_{t,r}-q_{min,r},\ q_{max,r}-q_{t,r})
    }{
    q_{max,r}-q_{min,r}+\epsilon
    } .
    \label{eq:joint_limit_margin}
\end{equation}
In addition to IK-based validation, each waypoint is required to satisfy a probabilistic reachability prior:
\begin{equation}
    \log p(x_i^t)\ge\tau_{qgmm},
    \qquad
    x_i^t=[p_x,p_y,p_z,q_x,q_y,q_z,q_w]^T .
    \label{eq:qgmm_threshold}
\end{equation}
Quaternion signs are canonicalized to a consistent hemisphere before QGMM training and evaluation to avoid the double-cover ambiguity of unit quaternions. The feasible set $\Freach$ is therefore defined by workspace support, QGMM likelihood, full-trajectory IK consistency, joint-limit margin, trajectory continuity, and Jacobian-conditioning constraints. In this work, $\Freach$ denotes a pre-execution kinematic feasibility set rather than a complete collision-free or contact-dynamic planning set.

\subsection{Semantic-Mask-Constrained Manipulability Objective}

For a reachable trajectory, the VLM-assisted and rule-validated semantic mask at frame $t$ is
\begin{equation}
    \Smask^t
    =
    [s_{T_x}^{t},s_{T_y}^{t},s_{T_z}^{t},
    s_{R_x}^{t},s_{R_y}^{t},s_{R_z}^{t}],
    \label{eq:smask_def}
\end{equation}
where larger values indicate more task-critical Cartesian components. The mask determines the allowable base-frame Cartesian perturbation:
\begin{equation}
    |\Delta \xi_k^t|
    \le
    b_k^{max}(1-s_k^t),
    \quad
    k\in\{T_x,T_y,T_z,R_x,R_y,R_z\}.
    \label{eq:smask_bound}
\end{equation}
Here, $\Delta\xi^t$ denotes a 6D Cartesian perturbation expressed in the robot base frame, and $b_k^{max}$ is the maximum perturbation bound for the $k$-th Cartesian component. The perturbed waypoint is represented by the left-multiplied SE(3) update
\begin{equation}
    \tilde{T}_{tcp}^{t}
    =
    \exp(\widehat{\Delta\xi^t})T_{tcp}^{t},
    \label{eq:se3_base_update}
\end{equation}
where $\widehat{(\cdot)}$ maps a 6D twist to the Lie algebra $\mathfrak{se}(3)$. The perturbed TCP trajectory is
\begin{equation}
    \tilde{\mathcal{T}}_{tcp}^{(i)}
    =
    \{\tilde{T}_{tcp}^{t}\}_{t=0}^{N}.
    \label{eq:perturbed_tcp_traj}
\end{equation}
Thus, task-critical axes are preserved more strictly, while semantically relaxable axes may be adjusted along the robot base-frame Cartesian directions to improve kinematic quality.

The velocity manipulability ellipsoid at frame $t$ is represented as an SPD matrix:
\begin{equation}
    \mathbf{M}_t
    =
    J_s(q_t)J_s(q_t)^T+\epsilon I_6,
    \qquad
    \mathbf{M}_t\in\SPD(6).
    \label{eq:spd_manip}
\end{equation}
Given a trajectory-conditioned target manipulability profile $\{\mathbf{M}_{c,t}\}_{t=0}^{N}$, the discrepancy between the current and target ellipsoids is measured by the affine-invariant Riemannian metric:
\begin{equation}
\begin{aligned}
    \dairm(\mathbf{M}_t,\mathbf{M}_{c,t})
    =
    \Big\|
    \log\big(
    \mathbf{M}_t^{-\frac{1}{2}}
    \mathbf{M}_{c,t}
    \mathbf{M}_t^{-\frac{1}{2}}
    \big)
    \Big\|_F .
\end{aligned}
\label{eq:airm}
\end{equation}
The refinement objective is compactly written as
\begin{equation}
\begin{aligned}
\min_{\{q_t,\Delta\xi^t\}}
\quad
&\sum_{t=0}^{N}
\Big[
\lambda_{spd}\dairm^2(\mathbf{M}_t,\mathbf{M}_{c,t})
+\lambda_{\sigma}\Gamma_{\sigma}(q_t)\\
&\qquad
+\lambda_{\kappa}\Gamma_{\kappa}(q_t)
+\lambda_l\Phi_{limit}(q_t)
+\lambda_s\Phi_{smooth}(q_t)
\Big]
\end{aligned}
\label{eq:q_refine_obj}
\end{equation}
subject to $\tilde{\mathcal{T}}_{tcp}^{(i)}\in\Freach$ and the S-Mask bounds in Eq.~(\ref{eq:smask_bound}). Here
\begin{equation}
    \Gamma_{\sigma}(q_t)
    =
    (\sigma_{\min}(J_s(q_t))+\epsilon)^{-1},
    \qquad
    \Gamma_{\kappa}(q_t)
    =
    \log\kappa(J_s(q_t)).
    \label{eq:singularity_terms}
\end{equation}
PhysV2A first exploits redundant joint-space motion while keeping the selected TCP trajectory fixed. S-Mask-constrained Cartesian relaxation is activated only when fixed-TCP redundancy refinement is insufficient.

\section{Method}

\subsection{Overview}
PhysV2A is organized around three method-level components: trajectory-conditioned grasp feasibility, hierarchical reachability-gated selection, and semantic-mask-constrained manipulability refinement. Given a recovered 6D object trajectory and RGB-D-generated 6-DoF grasp candidates, PhysV2A first constructs grasp-conditioned TCP trajectory hypotheses by preserving each grasp's rigid attachment to the moving object. It then rejects infeasible hypotheses and ranks the remaining candidates according to downstream execution suitability. Finally, the selected reachable trajectory is refined by first exploiting robot redundancy and then, only when necessary, applying S-Mask-constrained Cartesian relaxation. This hierarchy prevents the optimizer from attempting to repair unreachable trajectories and constrains trajectory refinement within semantically meaningful motion bounds.

\subsection{Trajectory-Conditioned Grasp Feasibility}
For every grasp candidate $G_i$, the recovered object trajectory is converted into a TCP trajectory using the rigid object-to-TCP attachment defined in Eqs.~(\ref{eq:object_to_tcp})--(\ref{eq:tcp_rigid_fusion}). This operation transforms one object-centric visual motion prior into multiple grasp-conditioned robot trajectory hypotheses,
\[
    \{\mathcal{T}_{tcp}^{(i)}\}_{i=1}^{K},
\]
where different grasps induce different downstream TCP trajectories even when the object motion is identical. Therefore, grasp feasibility is not evaluated only by local RGB-D confidence. Each grasp is assessed according to the complete TCP trajectory it induces after being coupled with the video-derived object motion, because a locally plausible grasp may still lead to an unreachable, discontinuous, or poorly conditioned robot trajectory.

\subsection{Hierarchical Reachability-Gated Selection}
Each grasp-conditioned hypothesis is evaluated by a two-stage reachability gate. The first stage performs hard feasibility screening, including workspace support, QGMM reachability thresholding, first-frame and terminal IK validation, sequential IK consistency, joint-limit validity, joint-space continuity, and singularity-safety checks. Candidates that fail any hard constraint are discarded before refinement. The second stage ranks the surviving candidates using a composite execution-suitability score,
\begin{align}
    S_i ={}& \alpha S_i^{grasp}+\beta S_i^{reach}+\gamma S_i^{ik} \notag\\
    &+\delta S_i^{margin}+\eta S_i^{cont}+\rho S_i^{manip},
    \label{eq:candidate_score}
\end{align}
where all terms are normalized to $[0,1]$, and a larger value indicates higher execution suitability. The weights are fixed across all tasks as $\alpha=0.20$, $\beta=0.20$, $\gamma=0.20$, $\delta=0.15$, $\eta=0.10$, and $\rho=0.15$. Here, $S_i^{grasp}$ denotes local grasp confidence, $S_i^{reach}$ measures QGMM workspace support, $S_i^{ik}$ measures waypoint-level IK consistency, $S_i^{margin}$ and $S_i^{cont}$ penalize joint-limit proximity and discontinuity, and $S_i^{manip}$ favors larger singular-value margins and lower condition numbers. The highest-scoring reachable trajectory is selected as
\begin{equation}
    \mathcal{T}_{tcp}^{*}
    =
    \arg\max_{\mathcal{T}_{tcp}^{(i)}\in\Freach} S_i,
    \label{eq:selected_tcp_traj}
\end{equation}
and is passed to the refinement stage.

\subsection{Execution-Suitability Scoring}
The composite score in Eq.~(\ref{eq:candidate_score}) is computed over the surviving candidate set $\mathcal{C}_{valid}$ after hard reachability screening. All scalar features are normalized scene-wise to $[0,1]$ over $\mathcal{C}_{valid}$, using increasing normalization for favorable quantities and decreasing normalization for penalty-like quantities. Specifically, $S_i^{reach}$ is computed from the mean QGMM log-likelihood along the fused TCP trajectory, $S_i^{ik}$ from the fraction of IK-valid waypoints, $S_i^{margin}$ from the minimum normalized joint-limit margin, $S_i^{cont}$ from the maximum normalized joint jump, and $S_i^{manip}$ from the worst-case minimum singular value and maximum condition number along the trajectory. The weights in Eq.~(\ref{eq:candidate_score}) were chosen heuristically to balance local grasp quality, reachability support, IK consistency, joint-limit safety, trajectory continuity, and manipulability conditioning, and were kept fixed for all tasks without task-specific tuning. Thus, the score ranks candidates by trajectory-level execution quality rather than by local grasp plausibility alone.

\subsection{Semantic Mask Generation}
After a reachable trajectory is selected, PhysV2A generates a semantic mask to specify which Cartesian components should be preserved or relaxed during local refinement. The VLM is used only as a semantic constraint generator, not as a trajectory generator, low-level controller, or learned continuous parameter estimator. Its inputs include the task instruction, representative key frames, and trajectory statistics computed from the recovered 6D object motion.

The VLM prompt is organized as a structured semantic-reasoning template rather than a free-form instruction. It contains the task instruction, sampled key-frame evidence, phase semantics, global and phase-wise trajectory statistics, rule hints, output schema, and validation constraints. The trajectory statistics include translation displacement, range, variance, path length, cumulative rotation, dominant rotation axis, and in-phase rotation variation. The prompt explicitly states that the object has already been grasped, so the departure phase should not be interpreted as a pre-grasp approach. It also requires the VLM to return a JSON-formatted phase-level mask with critical, soft, and relaxable axes, quaternion-statistics-based reasoning, and a relaxation-budget explanation.

The phase-level mask is written as
\[
\Smask^{p}=[s_{T_x}^{p},s_{T_y}^{p},s_{T_z}^{p},s_{R_x}^{p},s_{R_y}^{p},s_{R_z}^{p}],
\]
where $p$ denotes departure, transport, or target arrival. The three-phase split is used as a coarse temporal prior for the tabletop transfer and insertion tasks considered in this work, rather than as a universal segmentation rule for all manipulation tasks. The allowed values $\{1.0,0.8,0.5,0.3,0.1\}$ are expert-designed ordinal levels rather than learned continuous weights. They correspond to fixed, strongly preserved, moderately constrained, relaxable, and weakly constrained Cartesian components, respectively. Since the downstream correction budget is $a_k^t=1-s_k^t$, larger mask values impose stricter semantic preservation, whereas smaller values leave more freedom for IK continuity and manipulability improvement. The lower bound of 0.1 prevents any component from becoming completely unconstrained, and the same discrete scale is kept fixed across tasks for consistency.

The predicted mask is projected onto the discrete value set and validated by phase-specific rules. Departure must preserve at least one translational component so that the object leaves the source region; transport is allowed to be more flexible because it mainly preserves the motion trend; target arrival preserves final placement-related translation axes and only constrains rotations that are visually or kinematically task-critical. The validated phase-level mask is expanded to a per-frame mask, and its complementary correction budget is
\begin{equation}
    a_k^t = 1-s_k^t .
\end{equation}
During Cartesian relaxation, the base-frame perturbation is bounded by Eq.~(\ref{eq:smask_bound}). Thus, the S-Mask does not directly modify the trajectory; it defines the allowable semantic relaxation space for subsequent manipulability refinement.

\subsection{Semantic-Mask-Constrained Manipulability Refinement}
PhysV2A refines the selected trajectory in a conservative two-level manner. The first level keeps $\mathcal{T}_{tcp}^{*}$ fixed and exploits the redundant degree of freedom of the RM75 7-DoF manipulator. Candidate joint configurations are locally searched around the sequential IK solution, and updates are accepted only if they preserve the TCP pose within the predefined FK tolerance while improving the manipulability objective and maintaining joint-limit and continuity constraints. If fixed-TCP redundancy refinement satisfies the manipulability-safety criteria, the refined joint trajectory is directly accepted.

When fixed-TCP refinement is insufficient, PhysV2A activates S-Mask-constrained Cartesian relaxation. At each selected waypoint, translational and rotational perturbations are sampled only within the S-Mask bounds, reconnected through sequential IK, and evaluated at the full-trajectory level. An update is retained only if the complete trajectory remains reachable, IK-consistent, smooth, satisfies the S-Mask bounds, and better conditioned according to the SPD manipulability objective.

Overall, reachability defines the admissible trajectory set, the S-Mask defines the task-preserving relaxation space, and SPD manipulability provides the execution-quality objective within that constrained space. A target SPD profile is constructed from feasible joint configurations observed during full-trajectory IK validation and local redundancy search. For each waypoint, candidate configurations are ranked by larger $\sigma_{\min}(J_s)$ and smaller $\kappa(J_s)$, and the best-conditioned feasible ellipsoid is used as $\mathbf{M}_{c,t}$; if no better feasible configuration is available, the current ellipsoid is retained. This design improves trajectory conditioning without allowing the optimizer to distort task-critical Cartesian motion components.

\section{Experiments}

\subsection{Experimental Setup}
Experiments were conducted on an RM75 7-DoF manipulator equipped with a parallel gripper and an Intel RealSense RGB-D camera. The TCP was defined at the physical grasp point of the gripper, and all grasp poses, fused trajectories, and optimized trajectories were expressed in the robot base frame. For each scene, the upstream visual module provided a video-derived object 6D trajectory. The 6-DoF grasp candidates were generated using the off-the-shelf GraspGen framework \cite{murali2025graspgen} from the segmented object point cloud reconstructed from RGB-D observations. We retained the top 50 candidates after basic geometric filtering, including workspace-bound and gripper-approach validity checks. These candidates were used only as grasp proposals; each candidate was fused with the recovered object motion and evaluated by the reachability-gated grasp--trajectory screening module.

\begin{figure*}[!t]
    \centering
    \includegraphics[width=0.94\textwidth]{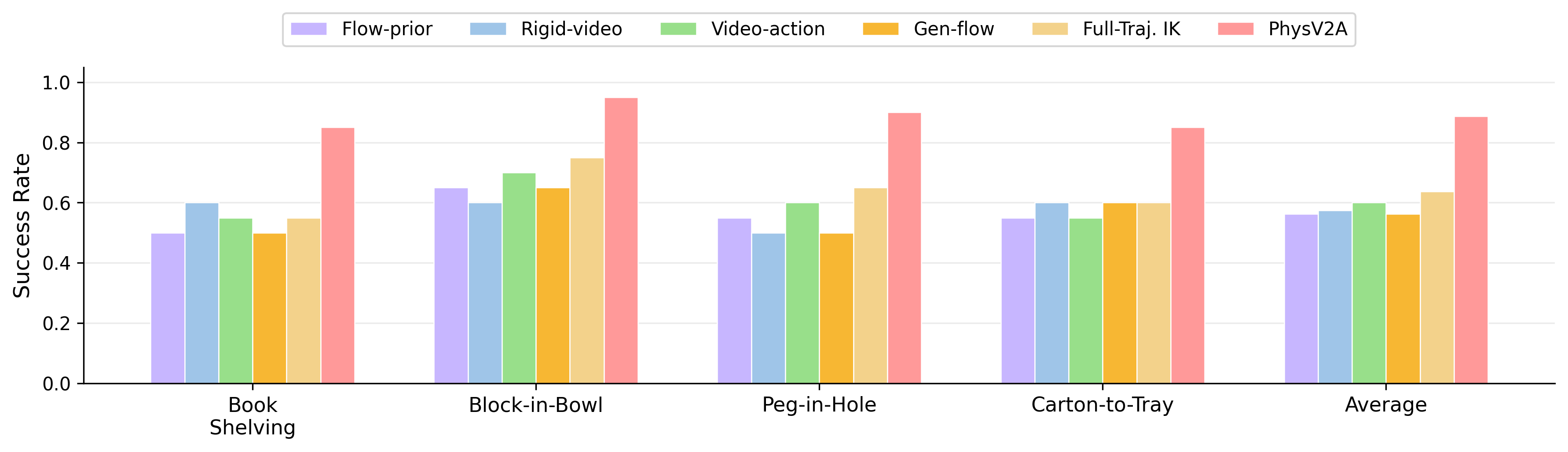}
    \caption{\textbf{Task success-rate comparison.} Flow-prior, rigid-video, video-action, and generated-flow retargeting are implemented as same-platform video-prior baselines using the same RM75 setup, RGB-D observations, grasp candidates, recovered object trajectories, and IK backend. PhysV2A achieves 71/80 successful trials and improves over Full-Traj. IK by 25.00 percentage points.}
    \label{fig:baseline_success}
    \vspace{-0.3em}
\end{figure*}

\begin{figure}[!t]
    \centering
    \includegraphics[width=0.95\columnwidth]{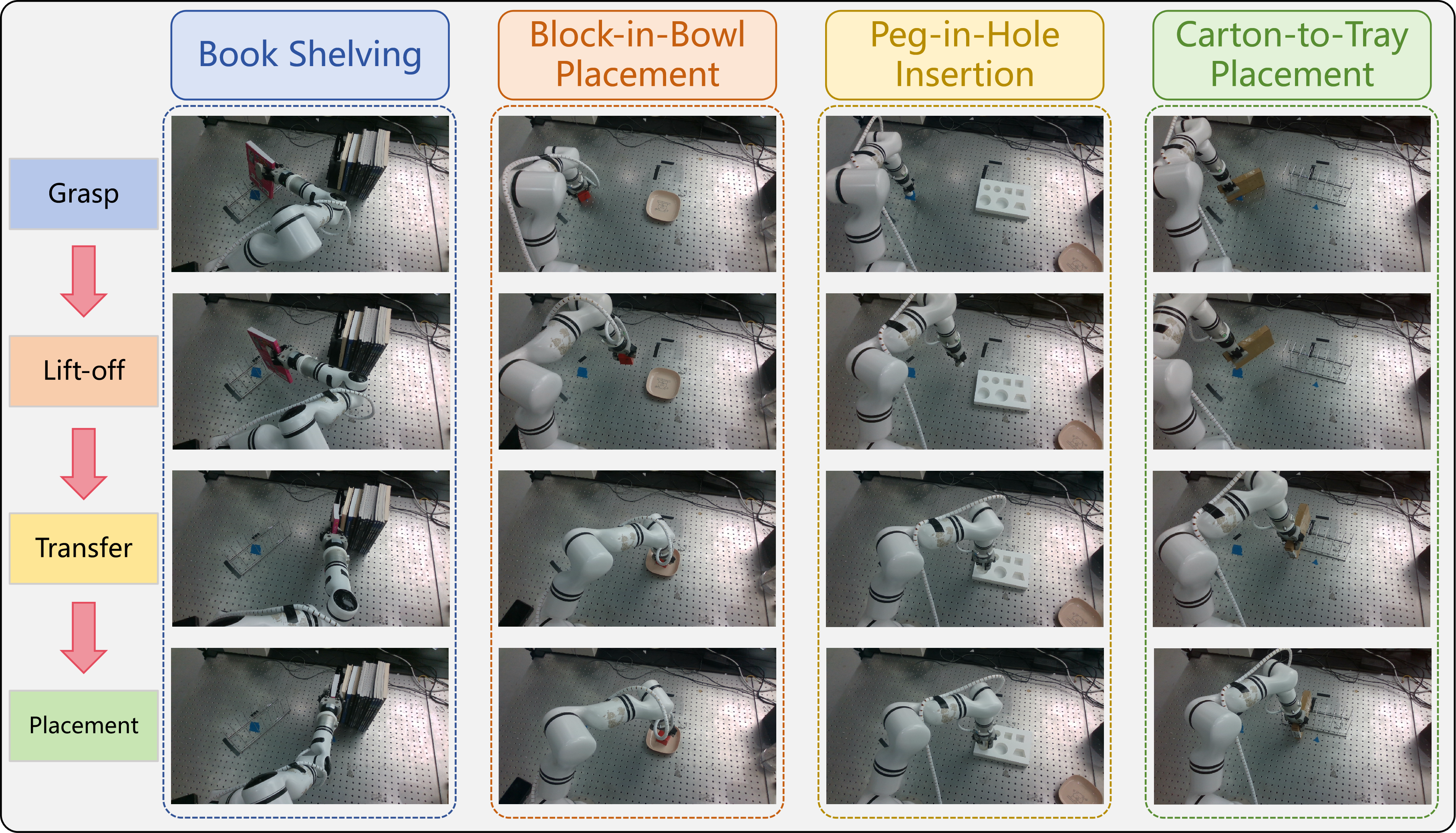}
    \caption{\textbf{Robot execution examples.} Representative execution phases across the four tabletop tasks.}
    \label{fig:real_execution}
    \vspace{-0.3em}
\end{figure}

Experiments were organized around four tabletop manipulation tasks, as illustrated in Fig.~\ref{fig:real_execution}: book shelving, block-in-bowl placement, peg-in-hole insertion, and carton-to-tray placement. These tasks cover long-horizon transfer, region-constrained placement, terminal pose precision, and object-geometry-dependent placement. Each method was evaluated for 20 independent trials on each task, resulting in 80 trials per method. The experiments evaluate whether grasp feasibility should be assessed at the trajectory level, whether hierarchical reachability-gated selection improves task success, and whether semantic-mask-constrained manipulability refinement improves kinematic quality without excessive trajectory distortion.

\subsection{Reachability Prior and Implementation Details}
The reachability gate uses an offline-trained QGMM model as a probabilistic prior over the RM75 TCP workspace. Joint configurations were sampled within URDF joint limits and mapped to TCP poses through forward kinematics using a Pinocchio-based RM75 model \cite{carpentier2019pinocchio}. The TCP frame was defined on \texttt{link\_7} with a 0.13~m offset along the local $z$-axis, and poses below the table plane were removed. Up to 200k table-filtered TCP poses represented by position and unit quaternion orientation were used to train a 24-component QGMM model. In all experiments, the QGMM threshold in Eq.~(\ref{eq:qgmm_threshold}) was set to $\tau_{qgmm}=-7.62$, empirically selected from the sampled reachable workspace distribution, and kept fixed across tasks without task-specific tuning. The likelihood is used only as a reachability prior; final executability is still determined jointly by workspace checks, endpoint IK, full-trajectory IK, joint-limit margin, continuity, and manipulability diagnostics.

All online variants used the same URDF-based kinematic backend for FK, IK, and Jacobian evaluation, with the previous-frame IK solution used as the rest configuration for the next frame. Unless otherwise stated, IK validation used $\epsilon_p=0.020$~m and $\epsilon_R=0.25$~rad for start and endpoint checks, while intermediate checks used relaxed tolerances of 0.050~m and 0.50~rad. All reported $\sigma_{\min}$, condition-number, and SPD AIRM metrics were computed from the scaled Jacobian $J_s$ in Eq.~(\ref{eq:scaled_jacobian}); in our setup, $l_0$ was set to the TCP offset length of 0.13~m and kept fixed across all methods and tasks. A candidate was treated as near-singular if $\sigma_{\min}<0.03$ or $\kappa>200$. The redundancy-first stage accepted a joint-only update only when the FK error to the fixed TCP waypoint was below 0.006~m and 0.08~rad. For S-Mask-constrained Cartesian relaxation, the maximum perturbation bound was $b^{max}=[10\,\mathrm{mm},10\,\mathrm{mm},10\,\mathrm{mm},\pi/36,\pi/36,\pi/36]$. Explicit self-collision and environment-collision checking were not included.

\subsection{Compared Baselines and Metrics}
To ensure a fair same-platform comparison, all methods share the same RM75 robot, RGB-D observations, segmented point cloud, 6-DoF grasp candidates, recovered object trajectory, grasp--trajectory fusion rule, kinematic backend, and IK initialization. Since the original video-conditioned methods differ in embodiment, sensing setup, policy representation, and training data, we implement representative same-platform retargeting baselines inspired by NovaFlow~\cite{li2025novaflow}, RIGVid~\cite{patel2025rigvid}, Gen2Act~\cite{bharadhwaj2025gen2act}, and Dream2Flow~\cite{dharmarajan2025dream2flow}, rather than claiming full reproduction of the original systems. These baselines convert visual motion or phase-level video priors into TCP trajectories through the same fusion and sequential IK pipeline, while excluding the proposed reachability gate, execution-suitability ranking, S-Mask, and SPD refinement.

Full-Traj. IK is used as an IK-only robot-centric baseline and is denoted as M1 in the internal ablation. It validates all 50 grasp-conditioned TCP trajectories by sequential IK and selects the IK-valid candidate with the lowest accumulated IK residual. For module-level analysis, M0 directly executes the trajectory induced by the highest-confidence grasp; M2 adds reachability-prior filtering, endpoint IK validation, joint-limit and continuity checks, and manipulability-aware scoring; and M3 is the full PhysV2A pipeline with VLM-assisted S-Mask generation, redundancy-first SPD manipulability refinement, and S-Mask-constrained Cartesian relaxation.

A trial is successful only when the robot grasps the object, executes the planned trajectory without IK failure, premature termination, or singularity-threshold violation, and completes the final placement or insertion. We report task, grasp, and execution success rates over the total number of trials, together with trajectory-quality metrics including SPD AIRM distance, minimum Jacobian singular value, maximum condition number, maximum joint jump, and bounded 6D trajectory deviation. Within each matched trial, all compared methods share the same observations, object trajectory, grasp candidates, robot model, and success criteria; only the enabled screening and refinement modules differ.

\begin{figure*}[!t]
    \centering
    \includegraphics[width=0.86\textwidth]{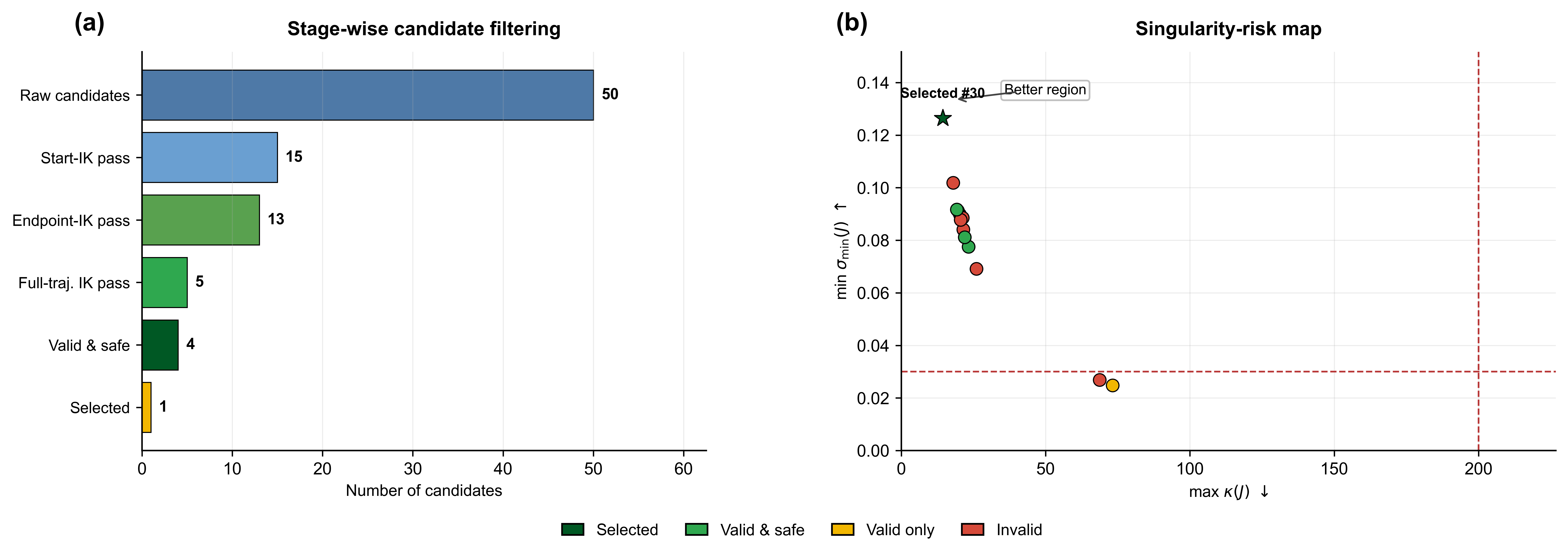}
    \caption{\textbf{Reachability-gated candidate selection.}
    (a) Stage-wise candidate filtering from initial grasp--trajectory candidates to the final selected candidate.
    (b) Singularity-risk map using minimum Jacobian singular value and maximum condition number.}
    \label{fig:reachability_selection}
    \vspace{-0.6em}
\end{figure*}

\subsection{Representative Baseline Success-Rate Comparison}
Figure~\ref{fig:baseline_success} compares PhysV2A with representative same-platform baselines across the four tabletop tasks. Flow-prior, rigid-video, video-action, and generated-flow retargeting obtain average task success rates of 56.25\%, 57.50\%, 60.00\%, and 56.25\%, respectively. Full-Traj. IK reaches 63.75\% by validating grasp-conditioned TCP trajectories with sequential IK. PhysV2A achieves the highest average success rate of 88.75\%, corresponding to 71 successful trials out of 80 and a 25.00 percentage-point improvement over the Full-Traj. IK baseline. This shows that the gain is not due to full-trajectory IK checking alone, but comes from trajectory-conditioned reachability evaluation, redundancy-first refinement, and S-Mask-constrained manipulability optimization.

\subsection{Main Real-Robot Results}
Table~\ref{tab:main_comparison} reports the multi-trial real-robot comparison among the internal same-platform variants. M0 succeeds in 45 out of 80 trials, M1, the IK-filtered selection variant, improves to 51, and M2 reaches 60 by introducing reachability-prior filtering, joint-limit and continuity checks, and manipulability-aware scoring. The full PhysV2A method achieves 71 successful trials, corresponding to eleven additional task successes over M2. At the task level, M3 achieves 17/20, 19/20, 18/20, and 17/20 successful trials on book shelving, block-in-bowl placement, peg-in-hole insertion, and carton-to-tray placement, respectively.

\begin{table}[!t]
\centering
\caption{Main real-robot comparison across four tabletop manipulation tasks. Each method was evaluated for 20 trials per task.}
\label{tab:main_comparison}
\scriptsize
\setlength{\tabcolsep}{3.0pt}
\resizebox{\columnwidth}{!}{%
\begin{tabular}{lccc}
\toprule
\textbf{Method} &
\textbf{Task SR} $\uparrow$ &
\textbf{Grasp SR} $\uparrow$ &
\textbf{Exec. SR} $\uparrow$ \\
\midrule
M0 Visual-motion-only
& 56.25\%
& 65.00\%
& 61.25\% \\
M1 IK-filtered selection
& 63.75\%
& 76.25\%
& 70.00\% \\
M2 Reachability-gated selection
& 75.00\%
& 87.50\%
& 80.00\% \\
\textbf{M3 Full PhysV2A}
& \textbf{88.75\%}
& \textbf{92.50\%}
& \textbf{90.00\%} \\
\bottomrule
\end{tabular}%
}
\vspace{-0.4em}
\end{table}

\begin{table}[!t]
\centering
\caption{Representative reachability-gated candidate screening results.}
\label{tab:reachability_screening}
\scriptsize
\setlength{\tabcolsep}{2.4pt}
\resizebox{\columnwidth}{!}{%
\begin{tabular}{lrrrrrrr}
\toprule
\textbf{Task} & \textbf{Raw} & \textbf{S-IK} & \textbf{E-IK} & \textbf{Full} & \textbf{Safe} & \textbf{Sel.} & \textbf{Score} \\
\midrule
T1 Book & 50 & 15 & 13 & 5 & 4 & \#30 & 0.9822 \\
T2 Bowl & 50 & 18 & 15 & 7 & 5 & \#17 & 0.9716 \\
T3 Peg & 50 & 12 & 9 & 4 & 3 & \#24 & 0.9638 \\
T4 Tray & 50 & 16 & 12 & 6 & 4 & \#11 & 0.9684 \\
\bottomrule
\end{tabular}%
}
\vspace{-0.3em}
\end{table}

\subsection{Reachability-Gated Selection Analysis}
Figure~\ref{fig:reachability_selection} and Table~\ref{tab:reachability_screening} summarize representative candidate screening results. Starting from 50 GraspGen-generated candidates per task, most high-confidence proposals are rejected by start-pose IK, endpoint IK, full-trajectory IK, or manipulability-safety checks. The selected candidate remains in a better-conditioned region with a larger singular-value margin and lower condition number, showing that local grasp confidence alone is insufficient for trajectory-level executability.

\subsection{S-Mask-Constrained Manipulability Refinement}
\begin{figure*}[!t]
    \centering
    \includegraphics[width=0.84\textwidth]{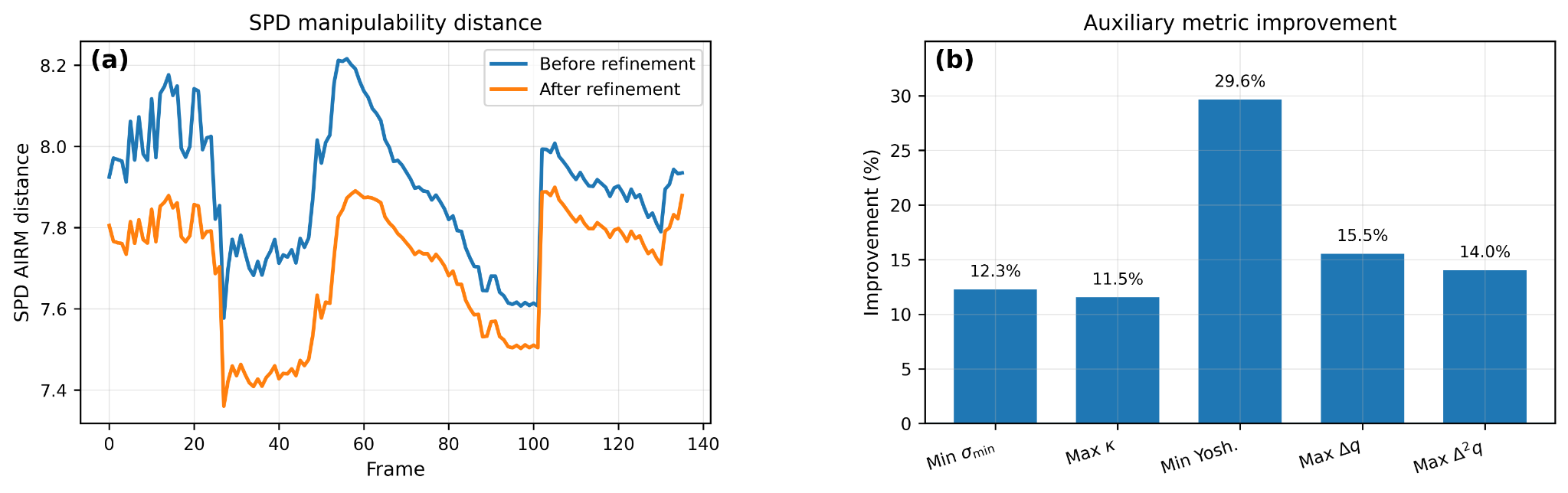}
    \caption{\textbf{S-Mask-constrained manipulability refinement.}
    (a) Frame-wise SPD AIRM distance before and after refinement.
    (b) Relative improvement of trajectory-level kinematic metrics with bounded Cartesian correction.}
    \label{fig:smask_manip_refinement}
    \vspace{-0.6em}
\end{figure*}

Figure~\ref{fig:smask_manip_refinement} presents the effect of S-Mask-constrained manipulability refinement. Compared with the reachability-gated trajectory, full PhysV2A reduces SPD AIRM distance and maximum condition number while increasing the minimum singular value, as summarized in Table~\ref{tab:trajectory_statistics}. The nonzero Cartesian deviations quantify bounded trajectory relaxation rather than independent failure indicators. This reflects a trade-off between pure manipulability improvement and task-semantic preservation: unrestricted Cartesian perturbation may further reduce manipulability cost, but it can also move terminal or phase-critical poses away from the intended task constraint. The S-Mask therefore acts as a semantic constraint boundary that limits optimization freedom on critical axes while allowing correction on less task-sensitive components.

\begin{table*}[!t]
\centering
\caption{Trajectory-level kinematic statistics over repeated trials. Cartesian deviations quantify the bounded adjustment introduced by refinement rather than independent failure metrics.}
\label{tab:trajectory_statistics}
\footnotesize
\setlength{\tabcolsep}{5.0pt}
\begin{tabular}{lcccccc}
\toprule
\textbf{Method} &
\textbf{SPD AIRM} $\downarrow$ &
\textbf{Min. $\sigma_{\min}$} $\uparrow$ &
\textbf{Max Cond.} $\downarrow$ &
\textbf{Max Joint Jump} $\downarrow$ &
\textbf{Pos. Adj.} &
\textbf{Rot. Adj.} \\
\midrule
M2 Reachability-gated selection
& $7.89 \pm 0.64$
& $0.106 \pm 0.018$
& $16.74 \pm 3.92$
& $0.184 \pm 0.041$
& $0.00 \pm 0.00$ mm
& $0.00 \pm 0.00^{\circ}$ \\
M3 Full PhysV2A
& $7.70 \pm 0.51$
& $0.119 \pm 0.016$
& $14.81 \pm 3.26$
& $0.155 \pm 0.035$
& $5.10 \pm 1.37$ mm
& $2.89 \pm 0.76^{\circ}$ \\
\bottomrule
\end{tabular}
\vspace{-0.3em}
\end{table*}

To evaluate S-Mask reliability, we report both repeat-run stability and human-label agreement in Table~\ref{tab:smask_stability}. For repeatability, a representative trajectory sample was evaluated across 20 independent VLM runs using the same structured prompt, key-frame evidence, trajectory statistics, allowed values, and rule-validation procedure. For human-label comparison, two human annotators independently labeled 20 trajectory samples, corresponding to 360 phase-axis mask entries. Human agreement was measured using exact agreement and quadratically weighted Cohen's $\kappa$. Disagreements were resolved through a discussion-based adjudication to form the human consensus label for each phase-axis entry. Critical axes are defined as phase-axis entries with mask values no smaller than 0.8.

\begin{table}[!t]
\centering
\caption{S-Mask repeatability on one representative trajectory and human-label agreement over 20 trajectories.}
\label{tab:smask_stability}
\scriptsize
\renewcommand{\arraystretch}{0.86}
\setlength{\tabcolsep}{2.5pt}
\resizebox{\columnwidth}{!}{%
\begin{tabular}{lc}
\toprule
\textbf{Metric} & \textbf{Value} \\
\midrule
\multicolumn{2}{l}{\textit{Repeat-run stability}} \\
Repeated runs on one trajectory (valid/requested) & 20/20 \\
Pairwise agreement (mean/min/max) & 0.884 / 0.722 / 1.000 \\
Cell-wise mode freq. (mean) & 0.893 \\
Consensus agreement (mean/min) & 0.897 / 0.778 \\
\midrule
\multicolumn{2}{l}{\textit{Human-label agreement}} \\
Trajectory samples / mask cells & 20 / 360 \\
Human agreement / weighted Cohen's $\kappa$ & 0.872 / 0.814 \\
VLM-human acc. (cell / one-step) & 0.846 / 0.912 \\
MAE / Critical-axis F1 & 0.071 / 0.858 \\
\bottomrule
\end{tabular}%
}
\renewcommand{\arraystretch}{0.94}
\vspace{-0.3em}
\end{table}

On the representative repeatability sample, the mean pairwise agreement and consensus agreement reached 0.884 and 0.897, respectively, indicating stable phase-axis mask generation under repeated prompting. The human annotations over the 20 trajectory samples showed good consistency, with an agreement of 0.872 and a weighted $\kappa$ of 0.814. Compared with the human consensus, the VLM achieved 0.846 cell-level accuracy and 0.912 one-step accuracy, where one-step accuracy treats predictions as correct if they exactly match the human label or fall within one adjacent ordinal level. The low mean absolute error of 0.071 and the critical-axis F1 score of 0.858 further indicate that the VLM-assisted S-Mask can identify task-relevant constrained degrees of freedom with reasonable reliability.

\subsection{Ablation Study}
Table~\ref{tab:ablation} reports the contribution of reachability gating, redundancy-aware refinement, semantic masking, and SPD manipulability refinement. The reachability gate provides the largest success-rate gain by removing kinematically infeasible grasp--trajectory pairs before refinement. Comparing the no-S-Mask variant with full PhysV2A further shows that better manipulability metrics alone do not necessarily lead to higher task success. Without semantic masking, the optimizer is allowed to apply less constrained Cartesian perturbations, which slightly improves the SPD AIRM distance and the minimum singular value but also introduces much larger translational and rotational deviations. These deviations can disturb task-critical terminal conditions such as placement position, insertion alignment, or object approach direction. Full PhysV2A yields slightly smaller pure manipulability gains but constrains relaxation within task-preserving axes, which explains its higher task success.

\begin{table*}[!t]
\centering
\caption{Ablation study of reachability, redundancy-first refinement, semantic masking, and manipulability refinement.}
\label{tab:ablation}
\scriptsize
\setlength{\tabcolsep}{3.0pt}
\resizebox{0.94\textwidth}{!}{%
\begin{tabular}{lccccccccc}
\toprule
\textbf{Variant} &
\textbf{Reach.} &
\textbf{Redun.} &
\textbf{S-Mask} &
\textbf{TCP Adj.} &
\textbf{SR} $\uparrow$ &
\textbf{SPD} $\downarrow$ &
\textbf{Min. $\sigma$} $\uparrow$ &
\textbf{Pos. Adj.} &
\textbf{Rot. Adj.} \\
\midrule
Visual-motion-only
& -- & -- & -- & --
& 56.25\%
& -- & 0.081
& 0.0 mm & 0.0$^\circ$ \\
+ IK-filtered selection
& IK-only & -- & -- & --
& 63.75\%
& -- & 0.087
& 0.0 mm & 0.0$^\circ$ \\
+ Reachability gate
& \checkmark & -- & -- & --
& 75.00\%
& 7.891 & 0.106
& 0.0 mm & 0.0$^\circ$ \\
+ Reachability + SPD, no S-Mask
& \checkmark & \checkmark & -- & \checkmark
& 80.00\%
& \textbf{7.642} & \textbf{0.121}
& 12.8 mm & 6.42$^\circ$ \\
\textbf{Full PhysV2A}
& \checkmark & \checkmark & \checkmark & \checkmark
& \textbf{88.75\%}
& 7.699 & 0.119
& \textbf{5.10 mm} & \textbf{2.89$^\circ$} \\
\bottomrule
\end{tabular}%
}
\end{table*}

\subsection{Qualitative Failure Analysis}
The remaining failures of the visual-motion-only baseline are mainly associated with incorrect grasp-region selection, unreachable grasp--trajectory pairs, trajectory IK failures, and low-manipulability configurations. IK-filtered selection reduces trajectory-level IK failures, but it still does not account for reachability-prior support, joint-limit margins, continuity, or manipulability-aware refinement. Reachability-gated selection substantially reduces workspace and trajectory-IK failures, while the full PhysV2A pipeline further reduces low-manipulability failures. The remaining failures are primarily associated with grasp execution uncertainty and final placement error, which require online perception or contact feedback beyond the current offline planning framework.

\section{Discussion}
The experimental results support the central claim that video-to-robot manipulation requires trajectory-conditioned feasibility reasoning rather than direct visual-prior retargeting. PhysV2A improves execution through three coupled design choices: evaluating grasp candidates by the TCP trajectories they induce, selecting candidates through a hierarchical reachability gate, and refining reachable trajectories only within semantic-mask-constrained Cartesian relaxation spaces. Although video-to-action methods such as NovaFlow~\cite{li2025novaflow}, RIGVid~\cite{patel2025rigvid}, Gen2Act~\cite{bharadhwaj2025gen2act}, and Dream2Flow~\cite{dharmarajan2025dream2flow} motivate the use of generated or video-derived visual priors, PhysV2A is designed as a model-agnostic kinematic feasibility-completion layer that can be integrated with any upstream module providing a 6D object trajectory and candidate grasps.

The SPD manipulability formulation is a geometry-aware execution-quality objective, not a closed-loop real-time manipulability controller. The no-S-Mask ablation shows that unrestricted Cartesian relaxation can improve manipulability indices while disturbing task-critical terminal conditions. In contrast, the S-Mask acts as a semantic boundary: it does not generate trajectories directly, but constrains where Cartesian relaxation is allowed. This explains why full PhysV2A can achieve higher task success while introducing smaller Cartesian deviations than the unrestricted variant.

Several limitations remain. PhysV2A depends on the quality of the upstream visual trajectory, assumes rigid object attachment after grasping, and focuses on pre-execution kinematic feasibility without explicit collision checking, contact dynamics, or closed-loop force control. The current S-Mask uses a coarse departure--transport--target-arrival phase template and an expert-designed ordinal scale, which are suitable for the tabletop transfer and insertion tasks studied here but may be insufficient for more complex long-horizon or contact-rich manipulation. For simple and repetitive tasks, manually designed task-specific masks may provide similar constraints; the advantage of the VLM interface is that it derives phase-wise semantic weights from instructions, visual evidence, and trajectory statistics without hand-specifying a new mask for every task. Future work will compare VLM-assisted masks, manual masks, and data-learned continuous masks, and will study adaptive phase segmentation, human-calibrated mask weights, collision-aware planning, tactile feedback, and contact-aware re-planning.

\section{Conclusion}
This paper presented PhysV2A, a reachability-gated and semantic-mask-constrained feasibility-completion framework for video-to-robot manipulation. The method is built on three key components: trajectory-conditioned grasp feasibility, hierarchical reachability-gated selection, and semantic-mask-constrained manipulability refinement. Experiments on four tabletop manipulation tasks show that PhysV2A outperforms representative video-prior and IK-only same-platform baselines, achieves an 88.75\% average task success rate, and provides eleven additional successful trials over the reachability-gated variant. The trajectory-level results further show improved manipulability, better Jacobian conditioning, and bounded Cartesian deviation, indicating that PhysV2A produces trajectories that are visually consistent with task intent, kinematically reachable, and semantically constrained.

\bibliographystyle{IEEEtran}
\bibliography{reference}

\begin{IEEEbiography}[{\includegraphics[width=1in,height=1.25in,clip,keepaspectratio]{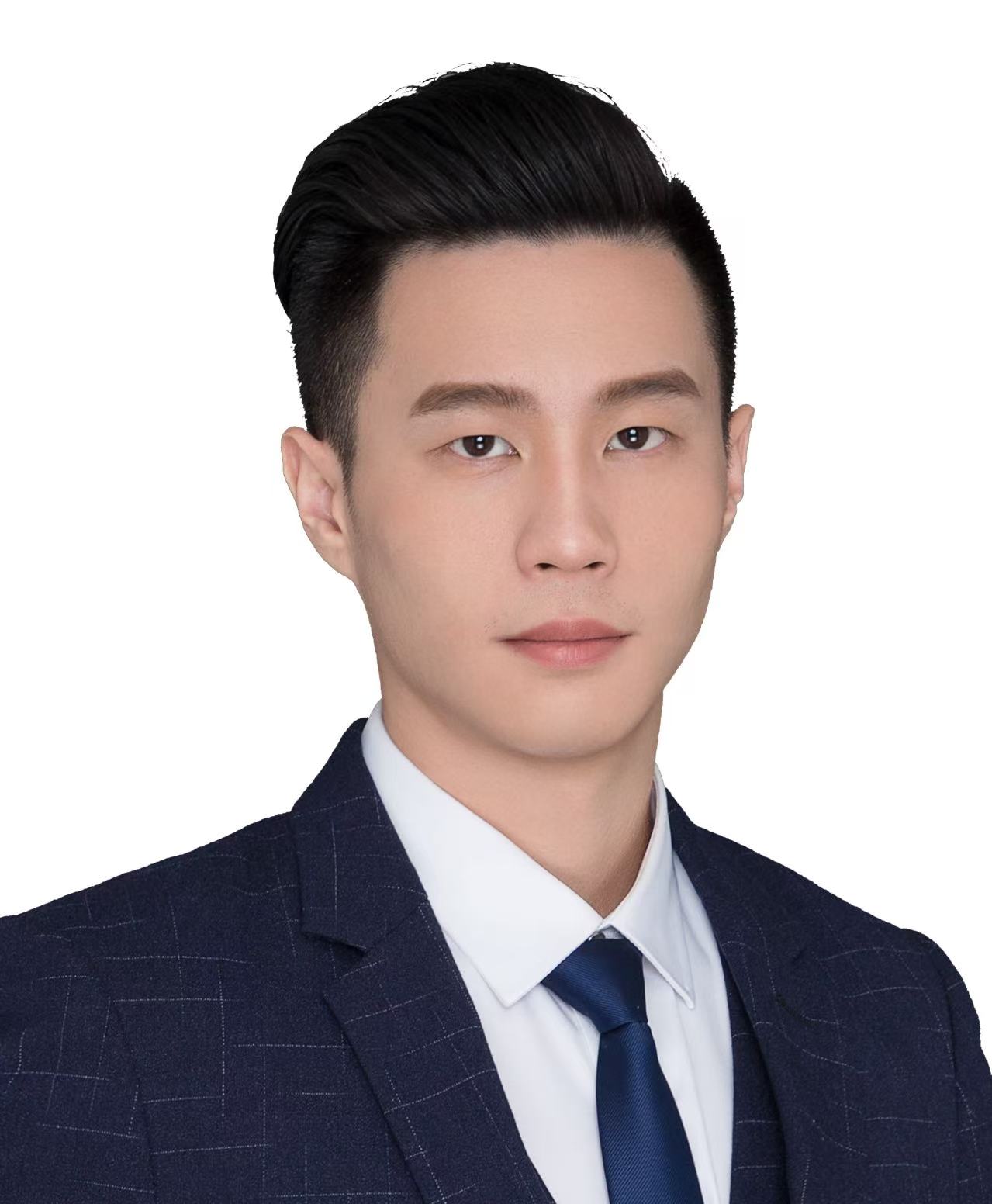}}]{Haohui Huang}
(Member, IEEE) received the B.S. and M.S. degrees from Guangdong University of Technology, Guangzhou, China, in 2011 and 2014, respectively, and the Ph.D. degree in control science and engineering from South China University of Technology, Guangzhou, China, in 2020. He completed postdoctoral training in human--robot interaction with Shanghai Jiao Tong University, Shanghai, China. Since 2023, he has been with the School of Automation, Guangdong University of Technology, Guangzhou, China. His research interests include robotics, compliant control, and human--robot interaction.
\end{IEEEbiography}

\begin{IEEEbiography}[{\includegraphics[width=1in,height=1.25in,clip,keepaspectratio]{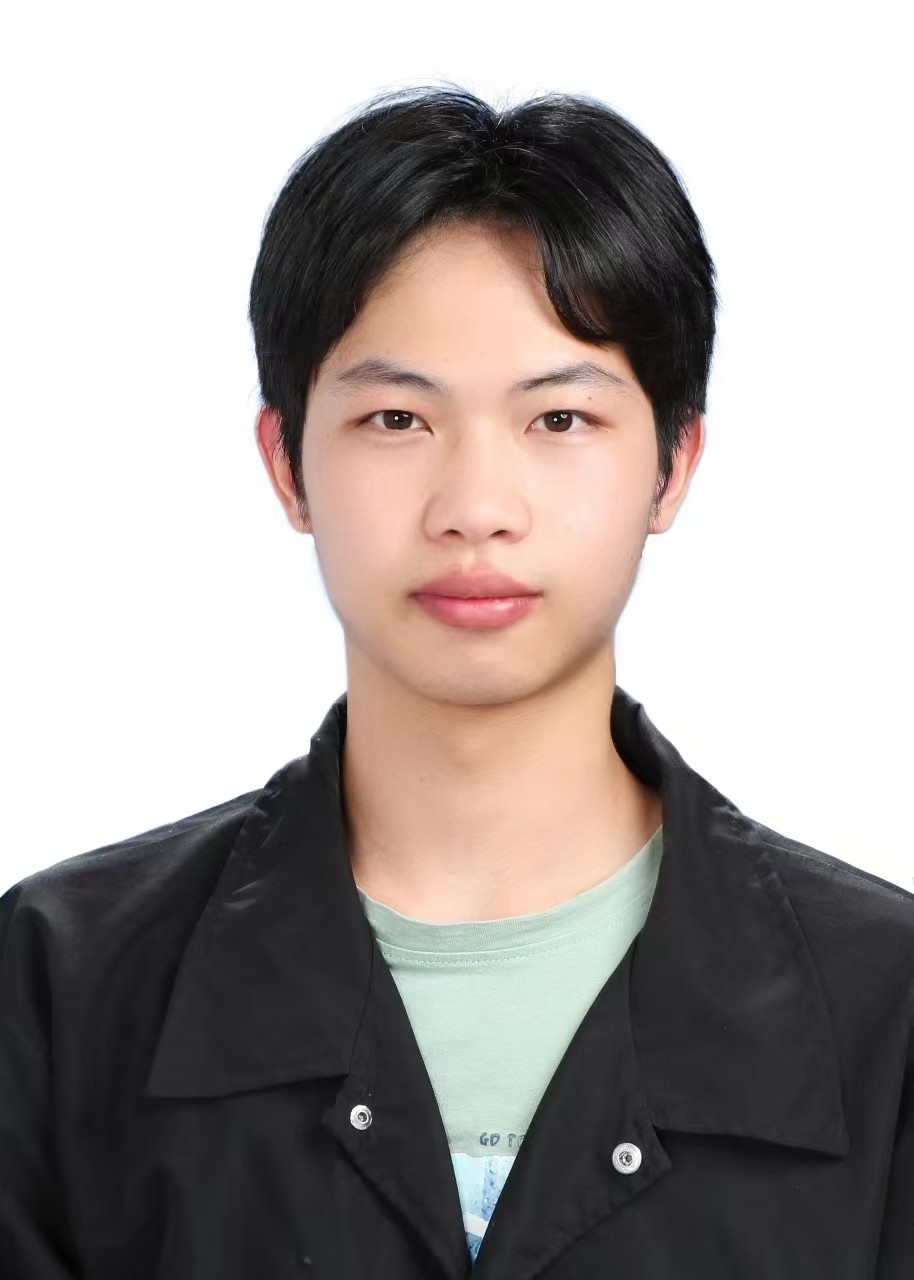}}]{Junda Duan}
received the B.S. degree in automation from Jiujiang University, Jiujiang, China, in 2024. He is currently a graduate student with the School of Automation, Guangdong University of Technology, Guangzhou, China. His research interests include robot imitation learning, motion control, and intelligent control.
\end{IEEEbiography}

\begin{IEEEbiography}[{\includegraphics[width=1in,height=1.25in,clip,keepaspectratio]{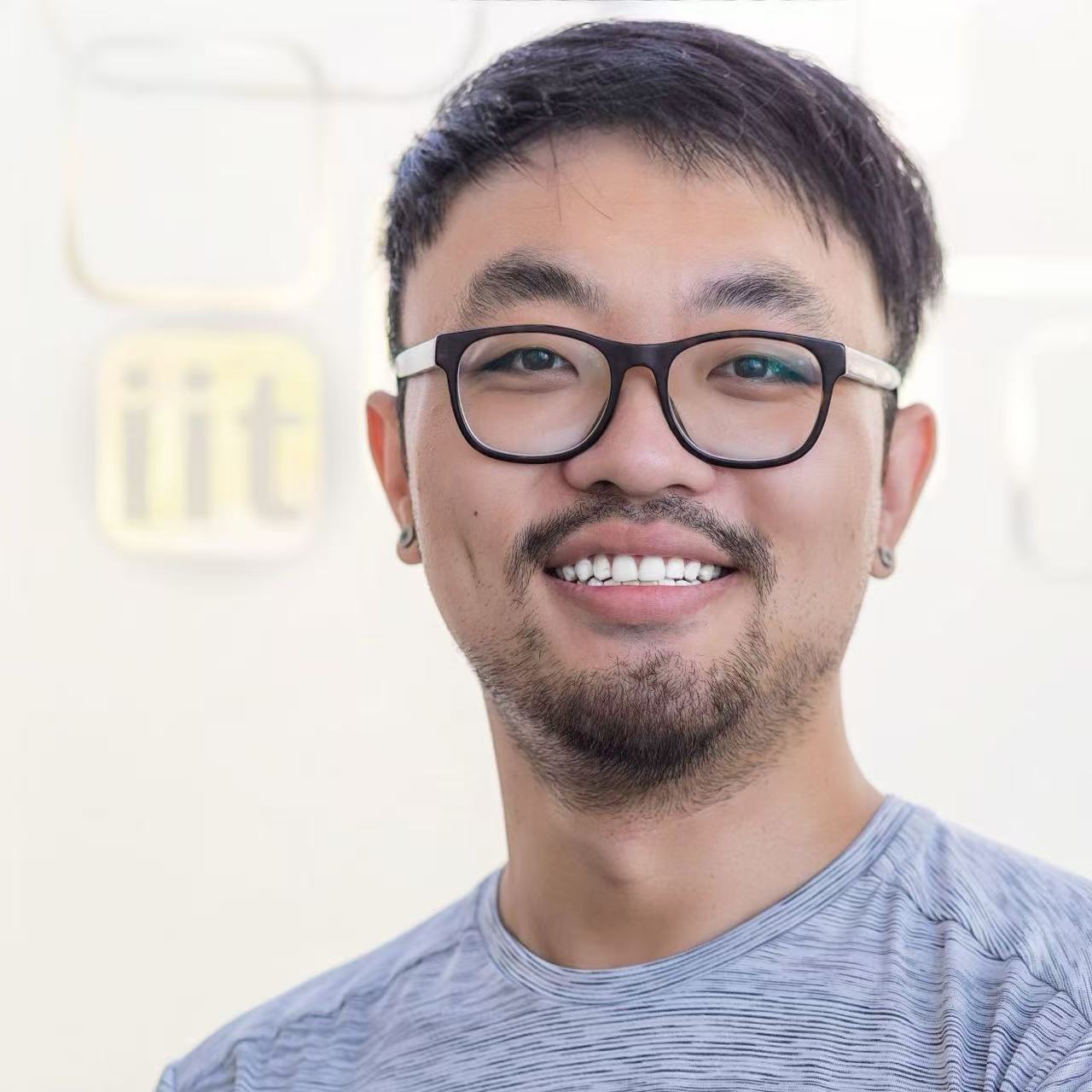}}]{Tao Teng}
(Member, IEEE) received the B.S. degree in automation and the M.S. degree in pattern recognition and intelligent systems from South China University of Technology, Guangzhou, China, in 2016 and 2019, respectively, and the Ph.D. degree in agri-food systems (robotics) from Universit\`a Cattolica del Sacro Cuore, in affiliation with the Istituto Italiano di Tecnologia, Italy, in 2023. He was a Postdoctoral Researcher in robotics with the Hong Kong Centre for Logistics Robotics and The Chinese University of Hong Kong. His research interests include human--robot interaction, robot learning, and control.
\end{IEEEbiography}

\begin{IEEEbiography}[{\includegraphics[width=1in,height=1.25in,clip,keepaspectratio]{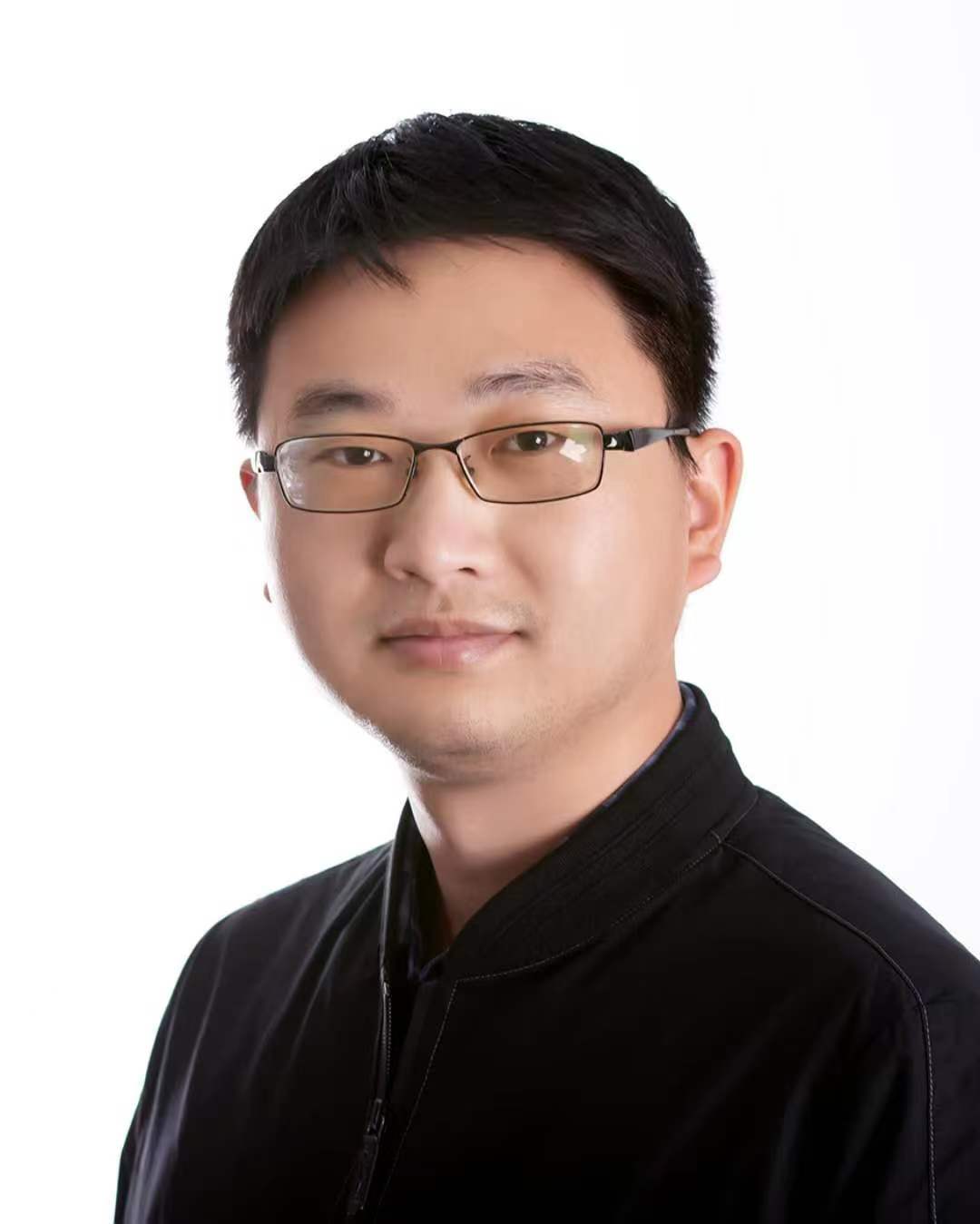}}]{Chenguang Yang}
(Fellow, IEEE) received the B.Eng. degree in measurement and control technique from Northwestern Polytechnical University, Xi'an, China, in 2005, and the Ph.D. degree in adaptive and neural network control from the National University of Singapore, Singapore, in 2010. He performed postdoctoral studies in human robotics at Imperial College London, London, U.K., from 2009 to 2010. Dr. Yang is currently a Professor with the Department of Computing, The Hong Kong Polytechnic University. He previously held professorships at the University of Liverpool, the University of the West of England (UWE Bristol), and South China University of Technology. He led the Robotics and Autonomous Systems Group at the University of Liverpool and the Robot Teleoperation Group at Bristol Robotics Laboratory. In addition to IEEE, he also holds fellowships with the Institution of Engineering and Technology (IET), the Institution of Mechanical Engineers (IMechE), the Asia-Pacific Artificial Intelligence Association (AAIA), and the British Computer Society (BCS). He is a member of the European Academy of Sciences and Arts (EASA) and the National Academy of Artificial Intelligence (NAAI). He was President of the Chinese Automation and Computing Society in the U.K. As the lead author, he won the IEEE Transactions on Robotics Best Paper Award in 2012 and the IEEE Transactions on Neural Networks and Learning Systems Outstanding Paper Award in 2022. His research interests include robot control and learning, human--robot interaction, and intelligent system design.
\end{IEEEbiography}

\end{document}